\documentclass[entropy,article,submit,moreauthors,LaTeX and dvi2pdf,10pt,a4paper]{mdpi}
\firstpage{1}
\articlenumber{x}
\doinum{10.3390/------}
\pubvolume{17}
\pubyear{2017}
\copyrightyear{2017}
\externaleditor{Academic Editors: name}
\history{Received: date; Accepted: date; Published: date}
\preto{\abstractkeywords}{\nolinenumbers}

\usepackage{microtype}
\usepackage{soul}
\usepackage{booktabs}
\usepackage{multirow}

\usepackage{algorithm}
\usepackage{algorithmicx}
\usepackage{algpseudocode}
\usepackage{multirow}
\usepackage[table,xcdraw]{xcolor}

\usepackage{microtype}
\usepackage{soul}
\usepackage{booktabs}
\usepackage{multirow}
\usepackage{rotating}
\usepackage{multicol}
\usepackage{lipsum}
\usepackage{makecell} 
\usepackage{amsmath}
\usepackage{amssymb}
\usepackage{epstopdf}
\usepackage{graphicx}
\usepackage{color}

\graphicspath{{figures/}}

 \theoremstyle{mdpi}
 \newcounter{thm}
 \newcounter{ex}
 \newcounter{re}
 \theoremstyle{mdpidefinition}

\Title{ Combination of interval-valued belief structures based on belief entropy }

\Author{Miao Qin $^{1}$, Yongchuan Tang $^{1}$*}

\AuthorNames{Miao Qin,  Yongchuan Tang}

\address{$^{1}$ {School of Big Data and Software Engineering, Chongqing University, Chongqing 401331 , China}; 20171979@cqu.edu.cn (M.Q.) }

\corres{Correspondence: tangyongchuan@cqu.edu.cn}

\abstract{This paper investigates the issues of combination and normalization of interval-valued belief structures within the
framework of Dempster-Shafer theory of evidence. Existing approaches are reviewed and thoroughly analyzed. The advantages and drawbacks of previous approach are presented. A new optimality approach  based on uncertainty measure is developed, where the problem of combining interval-valued belief structures degenerates into combining basic probability assignments. Numerical examples are provided to illustrate the rationality of the proposed approach.}

\keyword{Dempster--Shafer evidence theory; interval-valued belief structure; combination and normalization; decision making }

\begin{document}



%

\section{Introduction}
\label{Introduction}
First proposed by Dempster\cite{Dempster1967Upper}, and later refined and extended by Shafer\cite{shafer1976mathematical}, the Dempster-Shafer theory of evidence(DS theory or DST for short) has drawn a lot of attention. Its application involves a wide range of area including expert systems\cite{2001An}\cite{1988An}\cite{1996Measures}, information fusion\cite{2004Data}, pattern classfication\cite{2013Evidential}\cite{Reformat2008Building}\cite{2006Classification}, risk evaluation \cite{vandecasteele2016reasoning,zhang2017perceiving} \cite{2010The},  image recognition \cite{moghaddam2017toward}, classification\cite{Liu2017IEEE7747498,liu2019evidence} and data mining \cite{2013Maximum}  etc.

The original DS theory requires deterministic belie degrees and belief structures. However, in practical situations, evidence coming from multiple sources may be influenced by unexpected extraneous factors. The lack of information, linguistic ambiguity or vagueness and  cognitive bias all contribute to the uncertain evidence obtained in practical situations. For example, during risk assessment, expert may be unable to provide a precise assessment if he/she is not $100\%$ sure. In this case, an interval-valued belief structures instead of a deterministic belief structures seems more desirable under this circumstance.  In the group decision making, different experts may present different degrees of belief. By performing operations to average all evidence, it is possible to synthesize all information to elicit a precise evaluation, but this leads to potential loss of important information.
Therefore, the use of interval-valued
belief degree seems to be a rational option. This would preserve
the different belief degrees, thereby facilitating further discussion. Hence it is urgent to extend DS theory to interval-valued belief structures to deal with above problems.

Previous researchers made several attempts to extend the DS theory to solve the combination and normalization of interval-valued belief structures, such as Lee and Zhu\cite{1992An}, Denoeux\cite{1999Reasoning}\cite{2000Modeling}, Yager\cite{2001Dempster}, Wang et al.\cite{2007On} and Song et al.\cite{2014Combination}. However, each method has its own drawbacks, leaving the issue unresolved. The purpose of this paper is to reinvestigate the issues and to develop a new optimality approach based on uncertainty measure for combining and normalizing interval-valued belief structures.

 The rest of this paper is  organized as follows. In Section \ref{section2} we briefly introduce DS theory and concepts of interval-valued belief structures together with some basic arithmetic operations on intervals.
In Section \ref{section3}, we review and critically analyze the existing approaches for combining and normalizing interval-valued belief structures together with their drawbacks and advantages.
In Section \ref{section4}, a new optimality approach based on uncertainty measure is developed and the rationality and logic of the proposed method is also discussed.
Numerical examples are given in Section \ref{section5} to show implementation process and the performance of the proposed method.
Finally, the conclusion and a discussion of future work are given in Section \ref{section6}.

\section{Preliminaries}
	\label{section2}
The background material presented in this section includes 1) introduction of Dempster-Shafer Evidence theory(DS theory), 2)some basic knowledge of interval-valued belief structure, 3) operations on interval-valued belief structure and 4) Uncertainty measure in Dempster-Shafer theory and 5) transformation from basic probability assignment(BPA) to probability mass function(PMF)

\subsection{Introduction of Dempster-Shafer evidence theory}
\label{sec:2.1}
Dempster-Shafer evidence theory is widely applied in addressing uncertainties \cite{Zhou2017SELP,liu2019new}.

Some basic concepts are introduced as follows \cite{Dempster1967Upper,shafer1976mathematical}.

\textbf{Definition 1.} Let $X$ be a set of mutually exclusive and exhaustive events, denoted as:
\begin{equation}
X=\{\theta_{1},\theta_{2},\theta_{3}.....\theta_{|X|} \},
\end{equation}
where the set $X$ is named the frame of discernment(FOD). The power set of $X$ is defined as follows:
\begin{equation}
2^X=\{\text{\O},\{\theta_{1}\}......\{\theta_{|X|}\},\{\theta_{1},\theta_{2}\}......\{\theta_{1},\theta_{2}......\theta_{i}\}......|X|\},
\end{equation}

\textbf{Definition 2.} For a FOD $X=\{\theta_{1},\theta_{2},\theta_{3}.....\theta_{|X|} \}$, the mass function is a mapping $m$ from $2^{X}$ to $[0,1]$, defined as  follows:

\begin{equation}
m:2^{X}\rightarrow [0,1],
\end{equation}
which satisfies the following condition:
\begin{equation}
m\left \{ \emptyset \right \}=0\quad and\quad \sum_{A\in 2^{X}}m(A)=1.
\end{equation}
In DS theory, the mass function is also referred as basic probability assignment(BPA).
If a subset $a$ satisfies $a\in 2^{X}$ and $m(a)>0$, then $a$ is called the focal element of $m$.  If a BPA $m$ contains a focal element $X$ with belief 1, then the BPA $m$ is a vacuous BPA.

\textbf{Definition 3.} A Bayesian Belief structure is a belief structure on FOD $X$ whose focal elements are singleton of $X$. For example for belief structure $m(A)$ and FOD $X=\{\theta_{1},\theta_{2},\theta_{3}.....\theta_{|X|} \}$, if $m(A)=t$, $0\leq t\leq 1$ and $\left | A \right | =1$ where $\left | A \right | $ denotes the cardinality of $A$, then $m(A)$ is said to be a Bayesian Belief structure.

\textbf{Definition 4.}
The belief function and plausibility function associated with each BPA are defined as follows  respectively:
\begin{equation}
\begin{aligned}
Bel(A)&=\sum_{B\subseteq A}^{}m(B)  \\
Pl(A)&=\sum_{A\cap B\neq \varnothing }^{}m(B) \notag
\end{aligned}
\end{equation}
where $A$ and $B$ are subsets of FOD. The belief function $Bel$ can be interpreted as the lower bound of the probability to which target $A$ is supported and the plausibility function $Pl(A)$ can be seen as the higher bound. $Bel(A)$ and f$Pl(A)$ can be transformed to each other by the following equations:
\begin{equation}
Bel(A)=1-Pl(A) \notag
\end{equation}
and the inequality $Pl(A)\geq Bel(A)$ should always be satisfied.

\textbf{Definition 5.}
Assume there are two independent BPAs $m_{1}$ and $m_{2}$, the Dempster's rule of combination, which is denoted as $m=m_{1}\bigoplus m_{2}$, is defined as: follows\cite{Dempster1967Upper}:
\begin{equation}
\label{Equation5}
m(A)=\begin{cases}\frac{1}{1-K} \sum_{B\bigcap C=A}m_{1}(B)m_{2}(C),&\text{$A\neq\emptyset$,}
\\
0,&\text{$A=\emptyset$,}
\end{cases}
\end{equation}
where the $K$ is  is a normalization constant defined as follows:
\begin{equation}\label{Equation2}
K=\sum_{B\bigcap C=\emptyset}m_{1}(B)m_{2}(C)
\end{equation}

The normalization constant $K$ is assumed to be non-zero. if $K=0$, then $m_1$ and $m_2$ are in total-conflict and can not be combined using Dempster's rule. if $k=1$, $m_1$ and $m_2$ are non-interactive with each other, then $m_1$ and $m_2$ are non-conflicting. In order to combine multiple belief structures coming from different information sources, the following is applied:
\begin{equation}
m = m_1\oplus m_2\oplus ... \oplus m_k  \quad  k=1,2,3...N  \notag
\end{equation}
where $\oplus$ represents the combination operator.

The Dempster's rule of combination combines two BPAs in such a way that the new BPA represents a consensus of the contributing pieces of evidence. It also focus BPA on single set to decrease the uncertainty in the system based on the combination rule which can be useful in decision making process.

\subsection{ Interval-valued belief structure}
\label{sec:2.2}

\textbf{Definition 6.}
\label{Definition6}
For a FOD $X=\left \{\theta_{1},\theta_{2},\theta_{3}.....\theta_N  \right \}$, $F_1, F_2, F_3.....F_N$ be $N$ subsets of $X$, where $N$ denotes the cardinality of FOD $X$ and $N$ intervals $\left [ a_i, b_i \right ]$ with $0\leq a_i\leq b_i \leq 1 \quad(i=1, 2....N)$, an interval-valued belief structure on $X$ can be defined as follows:

(1) $a_i \leq m(F_i) \leq b_i$ where $0\leq a_i\leq b_i \leq 1$, $N$; \\
(2) $\sum_{i=1}^{N}a_i \leq 1$ and $\sum_{i=1}^{N}b_i \geq 1$; \\
(3) $m(\varnothing)  =0$

\textbf{Definition 7.}
\label{Definition7}
 Let $m$ be a interval-valued belief function with interval-valued probability masses $a_i \leq m(F_i) \leq b_i$ where $0\leq a_i\leq b_i \leq 1$, $(i=1, 2....N)$, a normalized interval-valued belief correspondent with $m$ must satisfy the following condition\cite{2007On}:
 \begin{equation}
 \label{Equation7}
 \sum_{i=1}^{N}b_i-(b_k-a_k)\geq 1\quad and\quad \sum_{i=1}^{N}a_i+(b_k-a_k)\leq 1 \quad for \quad \forall k\in \left \{ 1,2...N \right \}
 \end{equation}

 From the view of Wang et al.\cite{2007On}\cite{2006The},  there exist two kinds of non-normalized interval-valued belief function. The first kind
violates the condition of $\sum_{i=1}^{N} a_i \leq 1$ and $\sum_{i=1}^{N} b_i \geq  1 $. In this case, the interval-valued belief structure can be normalized by:
\begin{equation}
\begin{split}
\label{Equation8}
\bar{a_i} & = \frac{a_i}{a_i+\sum_{j=1,j \neq i}^{N}b_j}, i=1,2...N \\
\bar{b_i} & = \frac{b_i}{b_i+\sum_{j=1,j \neq i}^{N}a_j}, i=1,2...N
\end{split}
\end{equation}
Interval-valued belief structures that satisfy the condition $\sum_{i=1}^{N} a_i \leq 1$ and $\sum_{i=1}^{N} b_i \geq  1 $ but violates the condition in Definition \ref{Equation7} can be normalized by:
\begin{equation}
\begin{split}
\label{Equation9}
\bar{a_i} = max \left \{ a_i, 1-\sum_{j=1,j \neq i}^{N}b_j \right \} i=1,2...N \\
\bar{b_i} = min \left \{ b_i, 1-\sum_{j=1,j \neq i}^{N}a_j \right \} i=1,2...N
\end{split}
\end{equation}

\subsection{ Introduction of Intuitionistic fuzzy set}
\label{sec:2.3}
Intuitionistic fuzzy set (IFS), as one of the generations of the fuzzy set, was first proposed by Atanassov to deal with vagueness \cite{Atanassov1989Gargov}. The main advantage of the IFS is its property to cope with the uncertainty that may exist due to information impression. Because
it assigns to each element a membership degree and a non-membership degree, and thus, IFS constitutes an extension of Zadeh's
fuzzy set which only assigns to each element a membership degree. So IFS is regarded as a more effective way to deal with
vagueness than fuzzy set. Although Gau and Buehrer later presented vague set, it was pointed out by Bustince and Burillo that
the notion of vague sets was the same as that of IFS. Since its inception, IFS has been used in many areas, such as decision making
\cite{2005Multiattribute}\cite{Hong2000Multicriteria}\cite{2007Multicriteria} \cite{2007Multi} and pattern recognition\cite{2014A}\cite{2007Intuitionistic}. Here we briefly introduce some basic concepts of Intuitionistic fuzzy set.

\textbf{Definition 8.}
Let $X=\left \{ x_1, x_2...x_N \right \}$ be a universe of disclosure, then a fuzzy set $A$ in $X$ is defined as follows:
\begin{equation}
A=\left \{ \left \langle x, \mu _A(x) \right \rangle \mid x \in X \right \}
\end{equation}
where $\mu _A(x): X\rightarrow \left [ 0,1 \right ]$ is the membership degree.

\textbf{Definition 9.}
An IFS $A$ defined in $X=\left \{ x_1, x_2...x_N \right \}$  defined by Atanassov can be expressed as
\begin{equation}
A=\left \{ \left \langle x, \mu _A(x), \upsilon _A(x) \right \rangle \mid x \in X \right \}
\end{equation}
where $\mu _A(x): X\rightarrow \left [ 0,1 \right ]$ and $\upsilon _A(x): X\rightarrow \left [ 0,1 \right ]$ are membership degree
and non-membership degree, with the condition:
\begin{equation}
0 \leq \mu_A(x)+\upsilon_A(x) \leq 1
\end{equation}
The hesitancy degree of the element $x \in X$ to the set $A$ is defined as follows:
\begin{equation}
\pi _A(x)=1-\mu_A(x)-\upsilon_A(x)
\end{equation}
with the condition $\pi _A(x) \in \left [ 0,1 \right ], \forall x \in X $. Moreover, $\pi _A(x)$  is also called the intuitionistic index of $x$ to $A$. Greater $\pi _A(x)$ means more vagueness on $x$. Clearly when $\pi _A(x)=0$,  the IFS degenerates into an nromal fuzzy set.

\textbf{Definition 10.}
The following arithmetic operations on IFS are defined in order to perform combination or normalization on two IFS $A$ and $B$ on $X=\left \{ x_1, x_2...x_N \right \}$:

\begin{equation}
\begin{split}
&(1) A\subseteq B \, if \, \forall x\in X, \mu _A(x) \leq \mu _B(x) \, and \, \gamma_A(x) \leq \gamma_B(x). \\
&(2) A   =      B \, if \, \forall x\in X, \mu _A(x)  =   \mu _B(x) \, and \, \gamma_A(x) =    \gamma_B(x). \\
&(3) A^{C}= \left \{ \left \langle x, \mu _A(x), \upsilon _A(x) \right \rangle \mid x \in X \right \}, \, where \;  A^{C} \; is \; the \;  complementary  \; set  \; of \; A  \\
&(4) A \cap B = \left \{ \left \langle x, Min(\mu_A(x), \mu_B(x)), Max(\gamma_A(x), \gamma_B(x)) \right \rangle \mid x \in X \right \} \\
&(5) A \cup  B = \left \{ \left \langle x, Max(\mu_A(x), \mu_B(x)), Min(\gamma_A(x), \gamma_B(x)) \right \rangle \mid x \in X \right \} \\
&(6) A +  B = \left \{ \left \langle x, \mu_A(x)+\mu_B(x)-\mu_A(x)\cdot \mu_B(x), \gamma_A(x)\cdot \gamma_B(x) \right \rangle \mid x \in X \right \}  \\
&(7) A \cdot  B = \left \{ \left \langle x, \mu_A(x)\cdot \mu_B(x), \gamma_A(x)+\gamma_B(x)-\gamma_A(x)\cdot \gamma_B(x) \right \rangle \mid x \in X \right \}
\end{split}
\end{equation}
It can be proved that the above addition and multiplication rules are commutative and associative. However, the multiplication rule does not satisfy the distributive law, i.e., $A \cdot (B+C) \neq A \cdot B+ A \cdot C$ do not hold for all cases.

\subsection{Belief entropy in Dempster-Shafer evidence theory}
\label{sec:2.4}
Several method has been proposed in order to solve the problem of uncertainty measure in DS theory. Some previous methods are briefly summarized in Table \ref{Table1}.
\begin{table}[htbp]
  \centering
  \caption{Belief entropy in Dempster-Shafer evidence theory}
  \resizebox{\textwidth}{22mm}{
    \begin{tabular}{ccc}
    \toprule
    \multicolumn{1}{c}{\emph{Name}} & \multicolumn{1}{c}{\emph{Defination}} &  \\
    \midrule
      Dubois-Prade, Eq. \cite{dubois1987properties}  & $ H_{D}(m)=\sum_{A\subseteq X}m(a)log_{2}\left | A \right | $   \\
      Nguyen, Eq.  \cite{nguyen1987entropy} & $ H_{n}(m)=\sum_{A\subseteq X}m(A)log_2(\frac{1}{m(A)}) $   \\
 Klir-Ramer, Eq. \cite{mackay2003information} & $ H_{kr}(m)=-\sum_{A\subseteq X}m(A)log_{2}\sum_{B\subseteq X}m(B)\frac{\left | A\cap B \right |}{\left | B \right |}$   \\
Klir-Parviz, Eq. \cite{maeda1993uncertainty} & $H_{kp}(m)=-\sum_{A\subseteq X}m(A)log_{2}\sum_{B\subseteq X}m(B)\frac{\left | A\cap B \right |}{\left | A \right |}$   \\
Deng, Eq. \cite{deng2016deng} & $ H_{D}(m)=\sum_{A\subseteq X}m(A)log_{2}\frac{2^{\left | A \right |}-1}{m(A)} $   \\
Jirou sek-Shenoy, Eq. \cite{jirouvsek2018new} & $ H(m)=\sum_{x\subseteq  X}Pl\_P_m(x)log_2\frac{1}{Pl\_P_m(x)}+\sum_{A\subseteq X}m(A)log_2(\left | A \right |) $    \\
Yager, Eq. \cite{yager1983entropy} & $H_y(m)=-\sum_{A\subseteq X}m(A)log_2Pl(A) $    \\
Hohle, Eq. \cite{klir1992note} & $ H_o(m)=-\sum_{A\subseteq X}m(A)log_2Bel(A) $    \\
Pal {et al}, Eq.  \cite{pal1992uncertainty}& $ H_b(m)=\sum_{A\subseteq X}m(A)log_2(\frac{\left | A \right |}{m(A)}) $  \\
Qin, Eq. \cite{2020An}                                  & $ H_a(m)=\sum_{A\subseteq X }\frac{\left | A \right |}{\left |X  \right |}m(A)log_2(\left | A \right |)+\sum_{A\subseteq X}m(A)log_2(\frac{1}{m(A)}) $  \\
    \bottomrule
    \end{tabular}}%
  \label{Table1}%
\end{table}%

\section{Analysis of existing method to combine interval-valued belief structure }  
\label{section3}
The original D-S theory is originally designed to solve deterministic evidence in the form of BPA, not uncertain evidence with interval-valued belief structure. In interval-valued belief structure, the probability masses assigned to each focal elements can be vague and uncertain, which creates a lot of problems when trying to combine multiple uncertain evidence. Therefore, various methods intending to combine and normalize interval-valued belief structure have been proposed over the years. Here, we briefly introduce some of these methods and provide analysis based on its rationality.
\subsection{Lee and Zhu's method  }  
In order to make sure that the combination result of multiple interval-valued belief structure is closed, before Lee and Zhu introduce their approach in \cite{1992An}, they first defined two  basic arithmetic  operations on interval-valued belief structure:  summation  and multiplication. The definition is presented as follows:
\begin{equation}
\begin{split}
Generalized \; summation: \left [ a, b \right ]+\left [ c, d \right ]=\left [ u(a,c), u(b,d) \right ], \\
Generalized \;  multiplication: \left [ a, b \right ]\times \left [ c, d \right ]=\left [ i(a,c), i(b,d) \right ],
\end{split}
\end{equation}
where

\begin{equation}
\begin{split}
u(a,b)&= min\left [ 1,(a^{w}+b^{w})^{\frac{1}{w}} \right ]  \\
i(a,b)&= 1-min\left [ 1,((1-a)^{w}+(1-b)^{w})^{\frac{1}{w}} \right ]
\end{split}
\end{equation}
where $w\in  (0, \infty )$. Based on the the summation and multiplication on interval-valued belief structure, the combination of two pieces of interval evidence is as follows:

\begin{equation}
\left [ m_1\oplus m_2 \right ](C)=\sum_{A\cap B=C}^{}m_1(A)\times m_2(B),
\end{equation}
Like \cite{2007On} and \cite{2014Combination} pointed out, Lee and Zhu's approach have the following two drawbacks. One is the selection of
parameter $w$, which is highly subjective and arbitrary. Different choices of $w$ may lead to different combination results, which creates instability. The other is the use of non-normalized interval evidence both before and after combination. Here we demonstrate these two disadvantages by examining the following example. Consider the FOD $\Omega =\left \{ P, L, K \right \}$, two pieces of  uncertain evidence is presented as follows:

\textbf{Example 3.1}
\begin{equation}
\begin{split}
m_1(P)=\left [ 0.5, 0.8 \right ], m_1(LK)=\left [ 0.3, 0.4 \right ], m_1(\Omega)=\left [ 0.2, 0.5 \right ] \\
m_2(PL)=\left [ 0.4, 0.6 \right ], m_1(LK)=\left [ 0.3, 0.5 \right ], m_1(\Omega)=\left [ 0.3, 0.4 \right ]
\end{split}
\label{Example1}
\end{equation}
As mentioned in \cite{2014Combination}, the combination result varies when the value of parameter $w$ in Equation\ref{Equation8} changes. Certain selection of $w$ may even generate meaningless result.

\begin{table}[htbp]
  \centering
 \caption{Combination result for \textbf{Example 1}}
 \begin{tabular}{ccccccc}
  \toprule
     \multicolumn{1}{c}{w} &  \multicolumn{1}{c}{\emph{$\left \{ P \right \}$}} &  \multicolumn{1}{c}{\emph{$\left \{ L \right \}$}}&  \multicolumn{1}{c}{\emph{$\left \{ PL \right \}$}} &  \multicolumn{1}{c}{\emph{$\left \{ LK \right \}$}} & \multicolumn{1}{c}{$\left \{\Omega \right \}$} & \\
  \midrule
  1  &$\left [ 0,0.6 \right ]$ & 0  & $\left [ 0,0.1 \right ]$ & 0 & 0 \\
  2  &$\left [ 0.26,0.66 \right ]$ & $\left [ 0.08,0.28 \right ]$  & $\left [ 0,0.4 \right ]$ & $\left [ 0.01,0.40 \right ]$ &  $\left [ 0,0.22 \right ]$\\
  3  &$\left [ 0.34,0.64 \right ]$ & $\left [ 0.18,0.35 \right ]$  & $\left [ 0.10,0.43 \right ]$ & $\left [ 0.15,0.45 \right ]$ &  $\left [ 0.05,0.30 \right ]$\\
  4  &$\left [ 0.36,0.62 \right ]$ & $\left [ 0.22,0.37 \right ]$  & $\left [ 0.14,0.46 \right ]$ & $\left [ 0.20,0.46 \right ]$ &  $\left [ 0.10,0.34 \right ]$\\
  5  &$\left [ 0.38,0.61 \right ]$ & $\left [ 0.24,0.38 \right ]$  & $\left [ 0.17,0.47 \right ]$ & $\left [ 0.23,0.47 \right ]$ &  $\left [ 0.13,0.36 \right ]$\\
  \bottomrule
 \end{tabular}\label{Table2}
\end{table}

The combination of result presented in Table \ref{Table2} obviously shows that when $w=1$, the probability masses assigned to $\left \{ L\right \}$, $\left \{ LK \right \}$ and $\Omega$ are 0. It is impossible to make a decision based on this result. For $w=2,3,4,5$, as $w$ increases, the lower bound and upper bound for all focal elements increases  monotonically with $w$ as well. This trend seems trivial in Table \ref{Table2} but if $w$ is large enough, it is easy to conclude that the result will be distorted greatly. Recall Definition 6 and Definition 7, it is easy to see that these particular interval-valued belief structures do not satisfy the condition
mentioned in Definition 7, i.e they are not normalized. To meet the requirement $\sum_{i=1}^{N}a_i \leq 1$ and $\sum_{i=1}^{N}b_i \geq 1$ in Definition 6, The interval-valued belief structure transforms to the following:
\begin{equation}
\begin{split}
m_1(P)=0.5, m_1(LK)=0.3, m_1(\Omega)=0.2, \\
m_2(PL)=0.4, m_2(LK)=0.3, m_2(\Omega)=0.3,
\end{split}
\end{equation}
It has been pointed out in \cite{2007On} that  Lee and Zhu's method fails to capture this special case of deterministic evidence. The final result should be a combined  deterministic evidence instead of a new piece of interval evidence. Therefore without the consideration of normalization of interval evidence, the result of Lee and Zhu's method is questionable.

\subsection{Denoeux's method}
\label{sec:3.2}
Denoeux \cite{1999Reasoning}\cite{2000Modeling}systematically studied the issues of combination and normalization of interval evidence and he found that conventional interval arithmetic is not a good option to compute the combined probability masses of two pieces of interval evidence.He constructed the following pair of quadratic programming models to calculate the combined probability masses of two pieces of interval evidence:

\begin{equation}
\begin{split}
\label{Equation13}
Max/Min \, \left [ m_1 \oplus m_2 \right ](A)&=\sum_{B \cap C=A}^{} m_1(B)m_2(C) \\
s.t. \quad  \sum_{B\in F(m_1)}^{}m_1(B)&=1 \\
            \sum_{C\in F(m_2)}^{}m_2(C)&=1 \\
            {m_{1}}^{-}(B)\leq m_1(& B) \leq {m_{1}}^{+}(B) \\
            {m_{2}}^{-}(C)\leq m_2(& C) \leq {m_{2}}^{+}(C)
\end{split}
\end{equation}
where $m_1$ and $m_2$ are respectively two pieces of interval evidence with $B$ and $C$ as their focal elements, respectively. $m_1(B) \in \left [ {m_{1}}^{-}(B), {m_{1}}^{+}(B) \right ]$ and $m_2(C) \in \left [ {m_{2}}^{-}(C), {m_{2}}^{+}(C) \right ]$. $F(m_1)$ and $F(m_2)$  are respectively the sets of focal
elements of $m_1$ and $m_2$. $\left [ m_1 \oplus m_2 \right ](C)$ s the combined but non-normalized interval-valued probability mass of focal element $C$.

Denoeux also found that the combination of interval evidence has no property of associativity, i.e. $(m_1 \oplus m_2) \oplus m_3 \neq (m_1 \oplus m_3) \oplus m_2$
 in general. This lack of associativity is obviously a drawback and it makes the result of the combination of several interval-
valued belief structures dependent on the order in which they are combined. In order to overcome this drawback, he suggested combining all interval evidence in one step. After the combination of interval evidence in one step, the following normalization approach was developed to normalize all the combined but non-normalized interval-valued probability masses:
\begin{equation}
\begin{split}
m_{d}^{*-}(A)&= min\frac{m(A)}{1-m(\varnothing )} \\
             &=\frac{m^{-}(A)}{1-max\left [ m^{-}(\varnothing ), \sum_{B \neq A,B\neq \varnothing  }^{} m^{+}(B)-m^{-}(A) \right ]} \\      m_{d}^{*+}(A)&= max\frac{m(A)}{1-m(\varnothing )}  \\
             &=\frac{m^{+}(A)}{1-min\left [ m^{+}(\varnothing ), \sum_{B \neq A,B\neq \varnothing  }^{} m^{-}(B)-m^{+}(A) \right ]} \\
\end{split}
\end{equation}
where $m(A)\in\left [ m^{-}(A),m^{+}(A) \right ]$, $m(B) \in \left [ m^{-}(B),m^{+}(B) \right ]$, $m(\varnothing ) \in \left [ m^{-}(\varnothing),m^{+}(\varnothing) \right ]$ are non-normalized interval-valued probability masses, and $m_{d}^{*}(A) \in \left [ m_{d}^{*-}(A),m_{d}^{*+}(A) \right ]$ is the normalized interval-valued probability mass of $m(A) \in \left [ m^{-}(A),m^{+}(A) \right ]$. All the normalized interval-valued probability masses form a piece of interval-valued belief structure.

It is obvious that normalization and combination process of interval-valued belief structure are conducted separately and  optimized individually.It has been pointed out by Wang et al.\cite{2007On} that the resulting normalized interval-valued probability masses provided by Denoeux's approach are too wide to be true.

\subsection{Yager's method }
\label{sec:3.3}
Yager\cite{2001Dempster} also investigated the issue of normalization and combination process of interval-valued belief structure. However, \cite{2007On} pointed out that Yager's approach has two fundamental drawbacks. The first of them is the use of interval arithmetic in computing the combined probability masses of two pieces of interval evidence. The second drawback is the inability of this approach to normalize the combined interval-valued probability masses. Moreover, numerical example shown in \cite{2007On} that the upper bound of interval probability generated by the Yager's method may be less than the lower bound of this probability. In conclusion, it is impossible to construct a valid interval-valued belief structure via Yager's method which can be used in formal analysis.

\subsection{Wang et al. method}

Wang explored the issue  combination and normalization of interval-valued belief structures within the domain of D-S theory. They proposed a  optimality approach based on Denoex's previous work. Wang's approach is summarized below.

\textbf{Definition 8.}
They defined the interval-valued belief structure of $m_1 \oplus m_2$ as:
\begin{equation}
Max/Min \quad \left [ m_1 \oplus m_2 \right ](A)=\left\{
             \begin{array}{lr}
             \left [ (m_1 \oplus m_2)^{-}(A),(m_1 \oplus m_2)^{+}(A) \right ],\forall A\subseteq \Omega ,A \neq \varnothing &  \\
             0, A \neq \varnothing\\

             \end{array}
\right.
\end{equation}
where $ (m_1 \oplus m_2)^{-}(A)$ and $ (m_1 \oplus m_2)^{+}(A)$ denotes the minimum and the maximum value of the following optimization problems:

\begin{equation}
\begin{split}
\label{Equation23}
Max/Min \, \left [ m_1 \oplus m_2 \right ](A) &=\frac{\sum_{B \cap C=A}^{} m_1(B)m_2(C)
}{1-\sum_{B \cap C=\varnothing}^{} m_1(B)m_2(C)
} \\
s.t. \quad  \sum_{B\in F(m_1)}^{}m_1(B)&=1 \\
            \sum_{C\in F(m_2)}^{}m_2(C)&=1 \\
            {m_{1}}^{-}(B)\leq m_1(& B) \leq {m_{1}}^{+}(B) \\
            {m_{2}}^{-}(C)\leq m_2(& C) \leq {m_{2}}^{+}(C)
\end{split}
\end{equation}

The above optimization function can be extended to combine N interval-valued belief structure. The definition of $ m_1 \oplus m_2 \oplus...m_N $ is listed below:

\begin{equation}
Max/Min \quad \left [ m_1 \oplus m_2 \oplus... m_N \right ](A)=\left\{
             \begin{array}{lr}
             \left [ (m_1 \oplus m_2 \oplus... m_N)^{-}(A),(m_1 \oplus m_2\oplus... m_N)^{+}(A) \right ],\forall A\subseteq \Omega ,A \neq \varnothing &  \\
             0, A \neq \varnothing\\
             \end{array}
\right.
\end{equation}
where $ (m_1 \oplus m_2 \oplus...m_N)^{-}(A)$ and $ (m_1 \oplus m_2 \oplus...m_N)^{+}(A)$ denotes the minimum and the maximum value of the following optimization problems:

\begin{equation}
\begin{split}
Max/Min \quad \left [ m_1 \oplus m_2 \oplus...m_N \right ](A) &=    \frac{ \sum_{B_{j_1}^{1}\cap B_{j_2}^{2}...B_{j_N}^{N}=A}^{} m_1(B_{j_1}^{1}) m_2(B_{j_2}^{2}...m_N(B_{j_N}^{N}))}{1-\sum_{B_{j_1}^{1}\cap B_{j_2}^{2}...B_{j_N}^{N}=\varnothing}^{} m_1(B_{j_1}^{1}) m_2(B_{j_2}^{2}...m_N(B_{j_N}^{N}))} \\
s.t.  \quad \sum_{j=1}^{n_i}m_i(B_{j}^{i})&=1 \quad i=1,2...N \\
    m_{i}^{-}(m_{i}^{j})\leq & m(B_{i}^{j}) \leq m_{i}^{+}(B_{i}^{j}) \quad i=1,2...N \, \, j=1,2...n_i
\end{split}
\end{equation}

where  $m_1 ,m_2 , ... , m_N$ as N interval-valued belief structures, and $ m_{i}^{-}(m_{i}^{j})\leq  m(B_{i}^{j}) \leq m_{i}^{+}(B_{i}^{j}) \quad i=1,2...N \, \, , j=1,2...n_i $ as the probability masses of $m_i$, $B_{i}^{j}$ as the $j$th focal element of $m_i$.

Wang's method is based on Denoeux's previous work. Different from Equation\ref{Equation13}, each of the above models perform normalization and combination of interval-valued belief structures simultaneously instead of separately. According to Wang, this is necessary since it captures the true probability mass intervals of the combined focal elements. The good performance of Wang's method is illustrated by numerical examples given in \cite{2007On}. Although the optimality approach is logically sound, it has its limitations as well. First, obviously the non-associativity of Wang's method makes the combination result less convincible. Second, according to Wang, when combining three  or more pieces uncertain evidence, the normalization and combination process is conducted at the same time to obtain a precise solution in order to maintain associativity. Wang's method involves several nonlinear optimization models which brings up the computational complexity issue.

\subsection{Song et al. method}
\label{sec:3.5}
Song investigate the combination and normalization of interval-valued belief structures by utilizing pignistic transformation to obtain the so-called Bayesian belief structure. Song's method is briefly summarized as follows:

Suppose $ \Omega = \left \{ A_1, A_2, A_3 \right \}$ is the frame of discernment, after obtaining the Bayesian belief structure from pignistic transform,  An intuitionistic fuzzy set representing the belief structure m can be defined on X as:
\begin{equation}
M=\left \{ \left \langle A_i, \mu (A_i), \gamma (A_i) \right \rangle \mid A_i \in \Omega, i=1,2...N \right \}
\end{equation}

 where $m(A_i)=\left [ a_i, b_i \right ], i=1,2...N$ denote the probability masses assigned on each Bayesian belief structure and $\mu (A_i)$ and $1-\gamma (A_i)$  constitute the lower bound $a_i$ and upper bound $b_i$. The following relations can be easily obtained:
 \begin{equation}
 \begin{split}
 \label{Equation25}
 \mu (A_i) &=a_i  \\
 \gamma (A_i)&= 1-b_i \\
 \pi_(A_i) & = b_i - a_i
 \end{split}
 \end{equation}
According to  Dymova and Sevastjanov's interpretation of the triplet $ \mu (A_i),\gamma(A_i), \pi_(A_i)$ which, from the pattern recognition point of view, represents the answer to the question 'Does the unknown object $u$ belong to $A$', the FOD $M=\left \{ \left \{ Yes \right \},\left \{ No \right \},\left \{  Yes,No\right \} \right \}$ that corresponds with the triplet can be given. So, the membership degree $\mu_(A)$ can be treated as the probability mass of $u \in A$, i.e.,
as the focal element of the basic belief assignment function: $m^{A} (Yes)=\mu (A)$. Similarly, we can assume that $m^{A} (No)= \gamma(A) $. Since $\pi(A)$ is usually treated as the hesitation degree, a natural assumption is $m ^{A}(Yes,No) = \pi(A)$. Therefore, the IFS $M$ can elicit a belief structure as:
 \begin{equation}
\left\{\begin{matrix}
   \begin{split}
  &m^{A} (Yes)=\mu (A)\\
  &m^{A} (No)=\gamma (A) \\
  &m^{A} (Yes,No)=\pi (A)
   \end{split}
\end{matrix}\right.
 \end{equation}
 Given two IFS $M_1=\left \{ \left \langle A, \mu_1 (A) , \gamma_1(A)\right  \rangle \right \}$ and $M_2=\left \{ \left \langle A, \mu_2 (A) , \gamma_2(A)\right  \rangle \right \}$ on $\Omega= \left \{ A \right \} $, two belief structures with FOD $M=\left \{ \left \{ Yes \right \},\left \{ No \right \},\left \{  Yes,No\right \} \right \}$ can be given as:

\begin{equation}
\begin{split}
m_{1}^{A}(Yes)= \mu_1(A) \quad m_{1}^{A}(No)= \gamma_1(A) \quad m_{1}^{A}(Yes, No)= \pi_1(A)  \\
m_{2}^{A}(Yes)= \mu_2(A) \quad m_{2}^{A}(No)= \gamma_2(A) \quad m_{1}^{A}(Yes, No)= \pi_2(A)  \\
\end{split}
\end{equation}

Now we get two deterministic belief structure that can be combined using Dempster's combination rule:

\begin{equation}
\left\{\begin{matrix}
\begin{split}
&m_{1}^{A} \oplus m_{2}^{A}(Yes) =\frac{\mu_1(A)(1-\gamma_2(A))+\mu_2(A)\pi_1(A)}{1-\mu_1(A)\gamma_2(A)-\mu_2(A)\gamma_1(A)}  \\
&m_{1}^{A} \oplus m_{2}^{A}(No) = \frac{\gamma_1(A)(1-\mu_2(A))+\gamma_2(A)\pi_1(A)}{1-\mu_1(A)\gamma_2(A)-\mu_2(A)\gamma_1(A)}  \\
&m_{1}^{A} \oplus m_{2}^{A}(Yes,No) = \frac{\pi_1(A)\pi_2(A) }{1-\mu_1(A)\gamma_2(A)-\mu_2(A)\gamma_1(A)}
\end{split}
\end{matrix}\right.
\end{equation}
where $1-\mu_1(A)\gamma_2(A)-\mu_2(A)\gamma_1(A) \neq 1$.

Now based on above, given two IFS $M_1(\left \langle A_i,\mu_1(A_i), \gamma_1(A_i) \right \rangle \mid A \in \Omega )$ and $M_2(\left \langle A_i,\mu_2(A_i), \gamma_2(A_i) \right \rangle \mid A \in \Omega )$ on $\Omega =\left \{ A_1,A_2...A_N \right \}$, the combination operation is defined as follows:
\begin{equation}
\begin{split}
\label{Equation31}
M_1 \odot M_2 = \left \{ \left \langle A_i,\frac{\mu_1(A_i)(1-\gamma_2(A_i))+\mu_2(A_i)\pi_1(A_i)}{1-\mu_1(A_i)\gamma_2(A_i)-\mu_2(A_i)\gamma_1(A_i)}  ,\frac{\gamma_1(A)(1-\mu_2(A))+\gamma_2(A)\pi_1(A)}{1-\mu_1(A)\gamma_2(A)-\mu_2(A)\gamma_1(A)}\mid A=i-1,2...N \right \rangle \right \}
\end{split}
\end{equation}
where $\odot$ denotes the IFS combination operator.
Since Dempster's combination rule can be used to combine more than two belief structures, the above combination procedure of IFS can easily be extended to N IFS. With the commutativity and associativity of Dempster's rule, It is clear that Song's newly defined operation on IFS satisfies the following:
\begin{equation}
\begin{split}
&(1) M_1 \odot M_2 = M_2 \odot M_1  \\
&(2) (M_1 \odot M_2) \odot M_3 = M_1 \odot (M_2 \odot M_3)
\end{split}
\end{equation}
Based on Equation \ref{Equation25}, the resulting IFS \ref{Equation31} can be transformed to interval-valued belief function for each $A_i, i=1,2...N$. The transform result may not be normalized, therefore we  can normalize it using Equation \ref{Equation8} and Equation \ref{Equation9}.

Now Song's method can be briefly summarized as: 1) Obtain interval-valued belief structures. 2) Perform pignistic transform on the interval in order to get the so called  Bayesian belief structure.  3) Elicit the corresponding IFS for each element in FOD. 4) Combine all IFS based on the IFS combination operator. 5) Transform each resulting IFS back to interval belief structures. During step 1, 2 and 5,  the interval-valued belief structures may not necessarily be normalized, we can normalize it by Equation \ref{Equation8} and Equation \ref{Equation9}.

Pignistic transform \cite{dubois1982several}\cite{smets1989constructing} is  firstly proposed to transform BPA to PMF. Other methods such as maximum entropy credal set transform\cite{harmanec1994measuring}, and plausibility transform \cite{cobb2006plausibility}also aims at the same goal. In Song's method, by introducing IFS in the combination process, the original interval valued belief structures have to be transformed to Bayesian belief structures using pignistic transform. Now consider the following example:

\textbf{Example 3.2}: The FOD is given by $\Omega = \left \{ A,B,C \right \}$, The two pieces of interval evidence recorded below:
\begin{equation}
\begin{split}
&m_1(A)=\left [ 0.5,0.8 \right ],\quad m_1(BC)=\left [ 0.3,0.4 \right ] ,\quad m_1(\Omega)=\left [ 0.2,0.5 \right ] \\
&m_2(A)=\left [ 0.4,0.6 \right ],\quad m_2(BC)=\left [ 0.3,0.5 \right ] ,\quad m_2(\Omega)=\left [ 0.3,0.4 \right ]
\end{split}
\end{equation}
It is clear that the above interval-valued belief structures are not normalized, after normailizing it using Equation \ref{Equation8} and \ref{Equation9}, we get the following deterministic belief structures:
\begin{equation}
\begin{split}
&m_1(A)=\left [ 0.5 \right ],\quad m_1(BC)=\left [ 0.3 \right ] ,\quad m_1(\Omega)=\left [ 0.2\right ] \\
&m_2(A)=\left [ 0.4 \right ],\quad m_2(BC)=\left [ 0.3 \right ] ,\quad m_2(\Omega)=\left [ 0.3 \right ]
\end{split}
\end{equation}
After obtaining the normalized belief structures, we then utilize pignistic transform to elicit the Bayesian belief structures:
\begin{equation}
\begin{split}
m_{1}^{*}(A)=0.5666,\quad &m_{1}^{*}(B)=0.2167 ,\quad m_{1}^{*}(C)=0.2167 \\
m_{2}^{*}(A)=0.5,\quad   &m_{2}^{*}(B)=0.25 ,\quad m_{2}^{*}(C)=0.25
\end{split}
\end{equation}
By Equation \ref{Equation25}, we can get two IFS $M_1$ and $M_2$ respectively. representing $m_{1}^{*}$ and $m_{2}^{*}$ as:
\begin{equation}
\begin{split}
&M_1=\left \{ \left \langle A,0.5666,0.4334 \right \rangle,\left \langle B ,0.2167,0.7833 \right \rangle, \left \langle C,0.2167,0.7833 \right \rangle \right \} \\
&M_2=\left \{ \left \langle A,0.5,0.5 \right \rangle,\left \langle B ,0.25,0.75 \right \rangle, \left \langle C,0.25,0.75 \right \rangle \right \}
\end{split}
\end{equation}
Noted that in $M_2$, the target $A$ has IFS $\left \langle A,0.5,0.5 \right \rangle$. Given the relation between IFS and belief structure mentioned in Equation \ref{Equation25},if we combine $M_1$ and $M_2$ using the IFS combination operator and elicit the corresponding belief structure, we have:
\begin{equation}
\begin{split}
m_1 \oplus m_2 (A) = 0.5666=m_{1}^{*}(A)
\end{split}
\end{equation}
This example shows that in this case, the final combination result for the $A$ equals to the Bayesian belief structure obtained from original interval-valued belief structure, meaning that $m_2$ provide no information about $A$. The result is irrational and counterintuitive since in this special case of deterministic evidence, by utilizing pignistic transform to get the so called Bayesian belief structure, Song's method assigns the probability masses of $m_2(\Omega)=0.3$ equally to each elements in FOD, resulting in the lack of consideration of information provided in $m_2$ for target $A$. Since Song still consider the problem of combination and normalization of interval-valued belief structure within the domain of DS theory, pignistic transform, as the example presented above may cause the result of Song's method to be distorted.

Now consider another similar example. Only this time we try to investigate whether IFS is consistent with Dempster combination rule without the interference of pignistic transform.

\textbf{Example 3.3}
\begin{equation}
\begin{split}
&m_1(A)=\left [ 0.5,0.8 \right ],\quad m_1(B)=\left [ 0.3,0.4 \right ] ,\quad m_1(C)=\left [ 0.2,0.5 \right ] \\
&m_2(A)=\left [ 0.4,0.6 \right ],\quad m_2(B)=\left [ 0.3,0.5 \right ] ,\quad m_2(C)=\left [ 0.3,0.4 \right ]
\end{split}
\end{equation}
Similar to Example 3.2, after normalizing the above interval-valued belief structures we get two distinctive deterministic evidence. Clearly $m_1$ and $m_2$ are already Bayesian belief structures, so according to Song's method, we obtain the corresponding IFS as follows:
\begin{equation}
\begin{split}
&M_1=\left \{ \left \langle A,0.5,0.5 \right \rangle,\left \langle B ,0.3,0.7 \right \rangle, \left \langle C,0.2,0.8 \right \rangle \right \} \\
&M_2=\left \{ \left \langle A,0.4,0.6 \right \rangle,\left \langle B ,0.3,0.7 \right \rangle, \left \langle C,0.3,0.7 \right \rangle \right \}
\end{split}
\end{equation}
Combine $M_1$ and $M_2$ using Equation \ref{Equation31} we get:
\begin{equation}
\begin{split}
M_1 \odot M_2 =\left \{ \left \langle A,0.4,0.6 \right \rangle, \left \langle B,0.1552,0.8448 \right \rangle,\left \langle C,0.0968,0.9032 \right \rangle \right \}
\end{split}
\end{equation}
According to Equation \ref{Equation25}, the normalized result is listed below:
\begin{equation}
\begin{split}
m(A)=0.6143 \\
m(B)=0.2380 \\
m(C)=0.1485
\end{split}
\end{equation}
Now reexamine Example 3.3, it is obvious that $m_1$ and $m_2$ are deterministic evidence and can be combined using Dempster's combination rule, we can easily get the following based on Equation \ref{Equation5}:
\begin{equation}
\begin{split}
m_1 \oplus m_2(A)= 0.5714 \\
m_1 \oplus m_2(B)= 0.2571 \\
m_1 \oplus m_2(C)= 0.1714
\end{split}
\end{equation}
This example presents a inconsistency caused by introducing IFS in the combination and normalization process of interval-valued belief structures, which illustrate that in some cases Song's method may generate unstable result and affect the final decision making. In general, through Example 3.2 and Example 3.3, there are two drawbacks of Song's method: 1)Using pignistic transform to acquire Bayesian belief structures. 2)Constructing IFS model based on Bayesian belief structures. The examples show that in some circumstances Song's method generates unstable and irrational results.

\subsection{Summary}
The above review indicated that the issues of combination and normalization of interval evidence still needs further investigation so far. Each methods has its advantages and drawbacks. In the following section, we will develop a novel optimality approach for combining and normalizing interval-valued belief structures, unlike Wang's method which focuses on optimizing  the equation of Dempster's combination rule, we use uncertainty measure as a criteria to search for a pair of BPA that can represent a single interval-valued belief structure. We adopt the method proposed in Here KBS cite to normalize all interval-valued belief structures. It will be shown that the generated interval for each focal element can be regarded as the subset of the so-called true interval obtained by Wang et al.

\section{ A novel optimality approach for combining interval-valued belief structures}  
\label{section4}
Recall Wang's method, when combining and normalizing two interval-valued belief structures, the optimality approach selects a specific BPA by certain criteria,in this case the maximum and minimum value of Equation \ref{Equation23} with constraints, to degenerate the problem to combine BPAs using Dempster's combination rule. By following this logic, we introduce uncertainty measure to the combination and normalization process of interval-valued belief structures and try to elicit two BPAs from the original interval-valued belief structure with maximum and minimum entropy by using the uncertainty measure as the objective function. After obtaining a set of BPAs, it can be combined with BPAs acquired from other interval-valued belief structures to construct the interval for final results. The proposed method is breifly summarized below:

Suppose $m_1$ and $m_2$ are two interval-valued belief structures with interval-valued probability masses $m_{1}^{-}(A_i) \leq m_{1}^{}(A_i) \leq m_{1}^{+}(A_i)$ for $i=1...n_1$ and $m_{2}^{-}(B_j) \leq m_{2}^{}(B_j) \leq m_{2}^{+}(B_j)$ for $j=i...n_2$ respectively, with FOD $\Omega$. Based on different definition of uncertainty measure in Table \ref{Table1}, we first obtain the solution of the following  pair of optimization problems(Take Dubois-Prade, Eq as an example) for both $m_1$ and $m_2$, for $m_1$:
\begin{equation}
\begin{split}
\label{Equation43}
Max/Min \quad & H(m)=\sum_{i=1}^{n_1}m_1(A_i)log_2\left | A_i \right |   \\
s.t. \quad   & \sum_{i=1}^{n_1}m_1(A_i)=1 \\
             & {m_{1}}^{-}(A_i)\leq m_1(A_i) \leq {m_{1}}^{+}(A_i)
\end{split}
\end{equation}
for $m_2$:
\begin{equation}
\begin{split}
\label{Equation44}
Max/Min \quad & H(m)=\sum_{j=1}^{n_2}m_2(B_j)log_2\left | B_j \right |   \\
s.t. \quad   & \sum_{j=1}^{n_2}m_2(B_j)=1 \\
             & {m_{2}}^{-}(B_j)\leq m_2(B_j) \leq {m_{2}}^{+}(B_j)
\end{split}
\end{equation}
By transforming the Equation \ref{Equation43} and \ref{Equation44} to lagrange duality problem  , for $m_1$, we have a pair of BPA $m_{1}^{max}$ and $m_{1}^{min}$,  for $m_{1}^{max}$ being the solution of \emph{Max} Equation \ref{Equation43} and $m_{1}^{min}$ being the solution of \emph{Min} Equation \ref{Equation43}. From the stand point of uncertainty measure,  $m_{1}^{max}$ can be viewed as the BPA elicited from the original interval-valued belief structure $m_1$ that contains the most uncertainty, and  $m_{1}^{min}$ can be viewed as the BPA that contains the least uncertainty. Obviously, after optimization, the entropy of any other BPA that satisfy the constraints mentioned in Equation \ref{Equation43} belongs to the interval $\left [ H(m_{1}^{min}) ,H(m_{1}^{max})\right ]$. In this way, the original interval-valued belief structure $m_1$ is expressed as two BPAs $m_{1}^{max}$ and $m_{1}^{min}$ that represents the upper bond(the one containing the most uncertainty) and lower bond(the one containing the least uncertainty) of $m_1$, and the combination and normalization problem of interval-valued belief structures degenerate to the combination of BPAs in DS theory.

After performing the same procedure on $m_2$, now we have $m_{1}^{max}$, $m_{1}^{min}$ and $m_{2}^{max}$, $m_{2}^{min}$. Next we combine $m_{1}^{max}$ with $m_{2}^{max}$ and $m_{1}^{min}$ with $m_{2}^{min}$ using Dempster's combination rule mentioned in Equation \ref{Equation5}:
\begin{equation}
\begin{split}
(m_{1}^{max} \oplus m_{2}^{max})(C) =\frac{\sum_{A_i \cap B_j=C}^{}m_{1}^{max}(A_i)m_{2}^{max}(B_j)}{1-\sum_{A_i \cap B_j=\varnothing }^{}m_{1}^{max}(A_i)m_{2}^{max}(B_j)} \\
(m_{1}^{min} \oplus m_{2}^{min})(C) =\frac{\sum_{A_i \cap B_j=C}^{}m_{1}^{min}(A_i)m_{2}^{min}(B_j)}{1-\sum_{A_i \cap B_j=\varnothing }^{}m_{1}^{min}(A_i)m_{2}^{min}(B_j)} \\
\end{split}
\end{equation}

With these two BPAs, the resulting interval-valued belief structures $\dot{m}$ can be constructed based on the following:
\begin{equation}
\begin{split}
\label{Equation46}
\dot{m}(C) =\left [ Min((m_{1}^{max} \oplus m_{2}^{max})(C),(m_{1}^{min} \oplus m_{2}^{min})(C))  ,Max((m_{1}^{max} \oplus m_{2}^{max})(C),(m_{1}^{min} \oplus m_{2}^{min})(C)) \right ]
\end{split}
\end{equation}
with $A_i \cap B_j= C$ and $i=1...n_1 $, $j=1...n_2$. If the result is non-normalized, we can normalize it using Equation \ref{Equation8} and \ref{Equation9}.

The rationality of our method can be justified based on  interpretation of interval-valued belief structures from the perspective of information theory. What distinguishes interval-valued belief structures from BPAs is the uncertainty, or interval, of probability masses assigned to each focal elements. Our optimality approach views  the normalized interval-valued belief structures  as infinite number of BPAs that lies within the original interval. The question is how to determine a compact and equivalent form of BPA to fully represent the information contained in interval-valued belief structures. By introducing the concept of information entropy to this problem, the proposed method selects two BPAs as upper and lower bound of the original interval-valued belief structure based on "the amount of information" or "uncertainty" contained in BPAs. Apparently, the proposed method can be extended to combine N interval-valued belief functions since the proposed method degenerates the problem of combining interval-valued belief structures into combining BPAs. Because the combination process of the proposed method depends solely on  Dempster's combination rule, the property of commutativity and associativity can be easily achieved\cite{Dempster1967Upper}:
\begin{equation}
\begin{split}
 m_{1}^{max}\oplus(m_{2}^{max} \oplus m_{3}^{max})&=(m_{1}^{max}\oplus m_{2}^{max} ) \oplus m_{3}^{max} \\
  m_{1}^{max} \oplus m_{2}^{max} &= m_{2}^{max} \oplus m_{1}^{max}
\end{split}
\end{equation}
Similarly, commutativity and associativity can also be achieved if  $ m_{1}^{max}$ is replaced by $ m_{1}^{min}$.

In the next section, some numerical examples are presented in order to show the implementation process and the rationality and stability of the proposed method. The Figure 1 shows the flow chart to combine N nterval-valued belief structures on $\Omega = \left \{ A_1...A_N \right \}$.

\section{Numerical examples}
\label{section5}
First, we adopt the example presented in \cite{2014Combination} to elaborate the implementation process of the our optimality approach.

\textbf{Example 4:} Let $\Omega=\left \{ A_1,A_2,A_3 \right \}$ be the frame of discernment. Two interval-valued belief structures on X are defined as:
\begin{equation}
\begin{split}
&m_1(A_1)=\left [ 0.2,0.5 \right ] ,m_1(A_1 A_2)=\left [ 0.3,0.7 \right ], \\
&m_1(A_1 A_3)=\left [ 0,0.4 \right ],m_1(\Omega)=\left [ 0.1,0.5 \right ]  \\
&m_2(A_1)=\left [ 0.2,0.5 \right ] ,m_2(A_1 A_2)=\left [ 0.1,0.2 \right ], \\
&m_2(A_1 A_3)=\left [ 0.3,0.7 \right ],m_2(\Omega)=\left [ 0,0.4 \right ]
\end{split}
\end{equation}
It can be proven that these two interval-valued belief structures satisfy the conditions defined in Definition 6 and 7, i.e. they are
normalized. Here we take Pal et al, Eq.\cite{pal1992uncertainty} as the objective function for simplicity and optimize $m_1$, $m_2$ with:
\begin{equation}
\begin{split}
\label{Equation43}
Max/Min \quad & H_b(m)=\sum_{A\subseteq \Omega}m(A)log_2(\frac{\left | A \right |}{m(A)})  \\
s.t. \quad   & m_1(A_1)+m_1(A_1 A_2)+m_1(A_1 A_3)+m_1(\Omega)=1 \\
             & 0.2\leq m_1(A_1) \leq 0.5 \\
             & 0.3 \leq m_1(A_1 A_2) \leq 0.7 \\
             & 0 \leq m_1(A_1 A_3) \leq 0.4 \\
             & 0.1 \leq m_1(\Omega) \leq 0.5 \\
\end{split}
\end{equation}
and
\begin{equation}
\begin{split}
\label{Equation43}
Max/Min \quad & H_b(m)=\sum_{A\subseteq \Omega}m(A)log_2(\frac{\left | A \right |}{m(A)})  \\
s.t. \quad   & m_2(A_1)+m_2(A_1 A_2)+m_2(A_1 A_3)+m_2(\Omega)=1 \\
             & 0.2\leq m_2(A_1) \leq 0.5 \\
             & 0.1 \leq m_2(A_1 A_2) \leq 0.2 \\
             & 0.3 \leq m_2(A_1 A_3) \leq 0.7 \\
             & 0 \leq m_2(\Omega) \leq 0.4 \\
\end{split}
\end{equation}
After solving the above equations, we manage to obtain two sets of BPAs $m_{1}^{max}$, $m_{1}^{min}$ and $m_{2}^{max}$, $m_{2}^{min}$ that indicates the BPA with maximum entropy and minimum entropy for $m_1$ and $m_2$, respectively.
\begin{equation}
\begin{split}
&m_{1}^{max}(A_1)=0.2, m_{1}^{max}(A_1A_2)=0.3 \\
&m_{1}^{max}(A_1A_3)=0.2 ,m_{1}^{max}(\Omega)=0.3 \\
&m_{2}^{max}(A_1)=0.2, m_{2}^{max}(A_1A_2)=0.2 \\
&m_{2}^{max}(A_1A_3)=0.3 ,m_{2}^{max}(\Omega)=0.3
\end{split}
\end{equation}
and
\begin{equation}
\begin{split}
&m_{1}^{min}(A_1)=0.5, m_{1}^{min}(A_1A_2)=0.4 \\
&m_{1}^{min}(A_1A_3)=0 ,m_{1}^{min}(\Omega)=0.1 \\
&m_{2}^{min}(A_1)=0.5, m_{2}^{min}(A_1A_2)=0.1 \\
&m_{2}^{min}(A_1A_3)=0.4 ,m_{2}^{min}(\Omega)=0
\end{split}
\end{equation}
Given the relationship between the combination results of $m_{1}^{max} \oplus m_{2}^{max}$, $m_{1}^{min} \oplus m_{2}^{min}$ and the final interval-valued belief structure defined in Equation \ref{Equation46}. The result is listed below:
\begin{equation}
\begin{split}
\label{Equation53}
\dot{m}(A_1)& =\left [ 0.49,0.91 \right ], \dot{m}(A_1A_2)=\left [ 0.05,0.21 \right ] \\
\dot{m}(A_1 A_3)& =\left [ 0.04,0.21 \right ], \dot{m}(\Omega)=\left [ 0,0.09 \right ]
\end{split}
\end{equation}
Based on previous definition of normalized interval-valued belief structures mentioned in Definition 6 and 7, it is easy to validate that the resulting interval-valued belief structure is valid and normalized.

To compare the result of the proposed method with other results obtained earlier by other methods, the following example(adopted from \cite{2014Combination})is taken into consideration.

\textbf{Example 5:} Two normalized interval-valued belief structures on $\Omega=\left \{ A_1,A_2,A_3 \right \}$ are defined as:
\begin{equation}
\begin{split}
&m_1(A_1)=\left [ 0.2,0.4 \right ] ,m_1(A_2)=\left [ 0.3,0.5 \right ], \\
&m_1(A_3)=\left [ 0.1,0.3 \right ],m_1(\Omega)=\left [ 0,0.4 \right ]  \\
&m_2(A_1)=\left [ 0.3,0.4 \right ] ,m_2(A_2)=\left [ 0.1,0.2 \right ], \\
&m_2(A_3)=\left [ 0.2,0.3 \right ],m_2(\Omega)=\left [ 0.1,0.4 \right ]
\end{split}
\end{equation}
The final combination result can be given based on the implementation process in Example 4.  The combination results obtained by Lee
and Zhu's method \cite{1992An}, Denoeux's method\cite{1999Reasoning}, Wang's method\cite{2007On}, Yager's method \cite{2001Dempster} and  Song's method\cite{2014Combination} are shown in Table \ref{Table3}. The combination result of the proposed method using various uncertainty measure is presented in Table \ref{Table4}.
\begin{table}[htbp]
  \centering
 \caption{Combination result obtained by other methods for \emph{Example 5}( $w=3$ for Lee and Zhu's method)}
 \begin{tabular}{ccccccc}
  \toprule
     \multicolumn{1}{c}{} &  \multicolumn{1}{c}{\emph{Lee and Zhu's method}} &  \multicolumn{1}{c}{\emph{Denoeux's method}}&  \multicolumn{1}{c}{\emph{Yager's method}} &  \multicolumn{1}{c}{\emph{Wang's method}} & \multicolumn{1}{c}{\emph{Song's method}} & \\
  \midrule
  $m(A_1)$  &$\left [ 0.05,0.35 \right ]$ & $\left [ 0.13,0.73 \right ]$   & $\left [ 0.20,0.75 \right ]$ & $\left [ 0.22,0.55 \right ]$& $\left [ 0.36,0.52 \right ]$\\
  $m(A_2)$ &$\left [ 0,0.31 \right ]$ & $\left [ 0.12,0.67 \right ]$  & $\left [ 0.15,0.59 \right ]$ & $\left [ 0.19,0.48 \right ]$ &  $\left [ 0.24,0.39 \right ]$\\
  $m(A_3)$ &$\left [ 0,0.23 \right ]$ & $\left [ 0.05,0.56 \right ]$  & $\left [ 0.07,0.51 \right ]$ & $\left [ 0.08,0.39 \right ]$ &  $\left [ 0.18,0.32 \right ]$\\
  $m(\Omega)$  &$\left [ 0,0.24 \right ]$ & $\left [ 0,0.43 \right ]$  & $\left [ 0,0.25 \right ]$ & $\left [ 0,0.21 \right ]$ &  -\\

  \bottomrule
 \end{tabular}
 \label{Table3}
\end{table}
\begin{table}[htbp]
  \centering
 \caption{Combination result obtained by the proposed method base on different uncertainty measure for \emph{Example 5}}
 \begin{tabular}{ccccccc}
  \toprule
     \multicolumn{1}{c}{} &  \multicolumn{1}{c}{\emph{Dubois-Prade, Eq.}} &  \multicolumn{1}{c}{\emph{Nguyen, Eq.}}&  \multicolumn{1}{c}{\emph{Deng, Eq.}} &  \multicolumn{1}{c}{\emph{Pal et al, Eq.}} & \multicolumn{1}{c}{\emph{Qin, Eq.}} & \\
  \midrule
  $m(A_1)$  &$\left [ 0.3467,0.4274 \right ]$ & $\left [ 0.3234,0.5301 \right ]$   & $\left [ 0.3467,0.5128 \right ]$ & $\left [ 0.3382,0.5128 \right ]$& $\left [ 0.3382,0.5128 \right ]$\\
  $m(A_2)$ &$\left [ 0.2533,0.3333 \right ]$ & $\left [ 0.2962,0.3614 \right ]$  & $\left [ 0.2533,0.3846 \right ]$ & $\left [ 0.2690,0.3846  \right ]$ &  $\left [ 0.2690,0.3846 \right ]$\\
  $m(A_3)$ &$\left [ 0.1867,0.2393 \right ]$ & $\left [ 0.1084,0.2854\right ]$  & $\left [ 0.1026,0.1867 \right ]$ & $\left [ 0.1026,0.2006 \right ]$ &  $\left [ 0.1026,0.2006 \right ]$\\
  $m(\Omega)$  &$\left [ 0,0.2133 \right ]$ & $\left [ 0,0.0951 \right ]$  & $\left [ 0,0.2133 \right ]$ & $\left [ 0,0.1922 \right ]$ &  $\left [ 0,0.1922 \right ]$ \\
  \bottomrule
 \end{tabular}
 \label{Table4}
\end{table}
According to Equation \ref{Equation7} of Definition 7, all the combined results in Table \ref{Table3} and \ref{Table4} except Lee and Zhu's are normalized interval-valued belief structures. Because of the good performance of Wang's method in \cite{2006The} when applied to a cargo ship problem and the sound logic of the method, here we take Wang's global optimization solution as the true interval. From Table \ref{Table3}, apparently the solution for Lee and Zhu's method is different from the other methods. Lee and Zhu's every interval but $m(\Omega)$ is on the left of the true interval, with some overlaps. Therefore, the result of Lee and Zhu's method seems unreliable and counterintuitive in this case.

Considering the defect of Denoeux's method and Yager's method previously discussed in Section \ref{sec:3.2} and \ref{sec:3.3}, it is indeed that in Table \ref{Table3} Denoeux's method and Yager's method provide intervals that seems too wide that they include all true intervals obtained by Wang's method. The wide range of the final intervals render the combination result too uncertain to be used to make a valid and reliable decision. In this case, Denoeux's method and Yager's method are not applicable.

It is argued in \cite{2014Combination} that the center of the true interval is more plausible and reachable than both lower and upper bounds since bounds can be reached only in the worst case situation i.e. boundary condition, while the center value can come from many
different pieces of evidence elicited by the interval evidence. Song claims the interval concentrates nearer to the center is
more probable and more desirable. From the perspective of belief entropy, our optimality approach selects $m_{i}^{max},i=1,2...N$ with maximum entropy and $m_{i}^{min},i=1,2...N$ with minimum entropy to represent the original interval-valued belief structure. Intuitively, the boundry of the final result obtained by the proposed method can only be reached when all interval-valued belief structures are expressed as the following BPAs: $m_{i}^{max}$ or $m_{i}^{min}$ for $i=1,2...N$, logically the center of the resulting interval is indeed easier to reach. Therefore we agree with Song's interpretation on how to evaluate the result for combining and normalizing interval-valued belief structures. With this knowledge, now we can properly assess the efficiency and rationality of Song's method and the proposed method since, as shown in Table \ref{Table3} and \ref{Table4}, the results obtained by both method are all enclosed in Wang's true intervals.

Back to Table \ref{Table3}, Song's solution is indeed enclosed in the true interval, and the resulting intervals are more concentrated around the center of the true interval. Now recall in Section \ref{sec:3.5}, Song's method defines a new operation on the original interval-valued belief structures, using pignistic transform to obtain a so-called interval-valued Bayesian belief structure. Besides the drawbacks and risks of performing such operation, the final result obtained by Song's result can only be interval-valued belief structures with singleton focal elements. For instance, similar to \textbf{Example 5} , with the same FOD, we have:

\textbf{Example 6:}
\begin{equation}
\begin{split}
&m_1(A_1A_2)=\left [ 0.2,0.4 \right ] ,m_1(A_2A_3)=\left [ 0.3,0.5 \right ], \\
&m_1(A_1A_3)=\left [ 0.1,0.3 \right ],m_1(\Omega)=\left [ 0,0.4 \right ]  \\
&m_2(A_1A_2)=\left [ 0.3,0.4 \right ] ,m_2(A_2A_3)=\left [ 0.1,0.2 \right ], \\
&m_2(A_1A_3)=\left [ 0.2,0.3 \right ],m_2(\Omega)=\left [ 0.1,0.4 \right ]
\end{split}
\end{equation}
For \textbf{Example 6} , the final result of Song's method will look like: $m_(A_1)=...$, $m_(A_2)=...$, $m_(A_3)=...$, while others are like: $m_(A_1A_2)=...$, $m_(A_2A_3)=...$, $m_(A_1A_3)=...$, $m_(\Omega)$. It is clear that the two form of results can not be compared directly. Song  proposes to elicit the interval-valued Bayesian belief structures from other method's solution to make all results consistent and comparable. Since the nature of this operation on interval-valued belief structures has not been studied fully and  the inconsistency of pignistic  transform with the semantics of DS theory\cite{2017A}, It is hard to draw a valid and firm conclusion about the efficiency and rationality of Song's method.

As shown in Table \ref{Table4}, for different uncertainty measure, solutions of the proposed method are all enclosed in the true intervals. When using Dubois-Prade, Eq. as the method to measure uncertainty, the result presents intervals that lie closer to the center of the true intervals which provides more certain information than any other measures. Considering the drawbacks mentioned above and in Section \ref{section3} of other methods, the proposed method is obviously more desirable.


\section{Conclusion}
\label{section6}
In this paper we have mainly focused on addressing the issues of combination of interval-valued belief structures.
The normalization of interval-valued belief structure is carried out by Equation \ref{Equation8} and \ref{Equation9} proposed by Wang et al.\cite{2007On}. The existing approaches for combining and normalizing interval evidence are briefly introduced and thoroughly analyzed. The results obtained by these methods are either non-normalized, or irrational, or distorted. After analysis, we discover that Wang's method is logically sound and the combination result can provide true intervals. We propose an optimality approach based on uncertainty measure to degenerate the combination of interval-valued belief structures to combining representative BPAs. Because the proposed method transform interval-valued belief structures to BPAS and relies solely on Dempster's combination rule to combine BPAs,  commutativity and associativity can be easily achieved. The good performance of the proposed mehod has been illustrated by numerical examples. Compared to other methods, results obtained by the proposed method are more reasonable and superior than other solutions.


%
\vspace{6pt}


\newpage
\acknowledgments{The work is partially supported by
in part by the National Key Research and Development Project of China (Grant No.  2018YFF0214700),
and the National Natural Science Foundation of China (Grant No.  61672117).
}


\conflictofinterests{The authors declare no conflict of interest.
}

\bibliographystyle{mdpi}
\renewcommand\bibname{References}
\bibliography{bibfile}

%



\end{document}